# Modeling and Robust Attitude Controller Design for a Small Size Helicopter

Miaolei HE[*a,b], Jilin HE[*a],

[a] The State Key Laboratory of High Performance Complex Manufacturing, Central South University, Changsha, 410083, China
[b] Robotics Institute, Carnegie Mellon University, Pittsburgh, Pennsylvania 15213, USA
Corresponding author: Miaolei He(rchml@hotmail.com), Jilin He(hejilin@csu.edu.cn)

*Abstract*—This paper addresses the design and application controller for a small-size unmanned aerial vehicle (UAV). In this work, the main objective is to study the modeling and attitude controller design for a small size helicopter. Based on a non-simplified helicopter model, a new robust attitude control law, which is combined with a nonlinear control method and a model-free method, is proposed in this paper. Both wind gust and ground effect phenomena conditions are involved in this experiment and the result on a real helicopter platform demonstrates the effectiveness of the proposed control algorithm and robustness of its resultant controller.

## I. INTRODUCTION

The small unmanned autonomous helicopter has the characteristics of high maneuverability, small size, low cost and vertical taking off and landing[1],[2]. It has a wide application in industrial, agricultural, military fields, etc. There are many universities and research institutes have carried out relevant research on unmanned helicopter flight control[3]-[5].

Designing a robust flight controller is still a challenging task to control a helicopter safely and automatically in wind, especially conducted on an actual helicopter in the real world. There are also many other advanced control algorithms are used in literature, which includes the nonlinear robust control method [6],[7], model predictive control method [8], fuzzy control method [9], sliding mode control[10], backstepping method [3], etc. Among these methods, the H∞ technique is an ideal choice to attenuate the effects of environmental disturbances on the vehicle system performance. Basar, et al. in [11], describes H∞ technique as a zero-sum game problem, the goal of the controller is to find a solution to the game, which means solving the problem Riccati equation for the nonlinear system. Guiherme V.Raffo, et al. in [12] present a predictive and robust H∞ control strategy for a helicopter, which was combined with model predictive control to realize rotational movement stabilizing and trajectory tracking.

It's hard to build an accurate model for the helicopter, manual measuring method does not consider frequency response scenario. Thus, the system identifying method in [13] is needed to improve the accuracy of the model. Actually, it is impossible to collect all kinds of data in every case, especially during landing and take-off periods. Both wind gust and ground effect phenomena conditions are involved in this scenery. In order to overcome this problem, a model-free method is introduced to provide attitude reference information in the attitude controller.

## II. HELICOPTER DYNAMIC MODEL

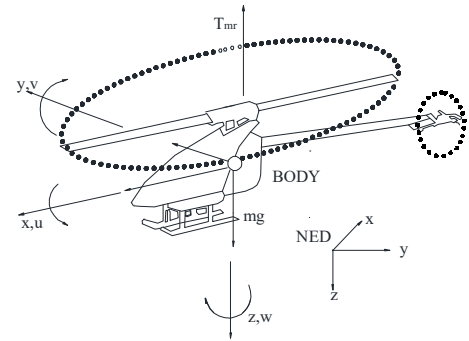

Fig. 1. the Schematic of the helicopter.

The dynamic model consists of four parts: the body kinematic characteristics, the body dynamics, main rotor dynamic and yaw channel characteristics.

### A. The Body Kinematic Characteristics

The kinematic model consists of translation motion and rotation motion, based on assuming that, the word north-east-down (NED) is an inertial system. Its direction and the direction of the body coordinate system are both pointing at the same direction, as shown in Fig.1. The equations of translation motion are as below:

$$\begin{cases} h = -P_{ned\_z} \\ \begin{bmatrix} \dot{P}_{ned\_x} & \dot{P}_{ned\_y} & \dot{P}_{ned\_z} \end{bmatrix}^T = R_B \begin{bmatrix} u & v & w \end{bmatrix}^T \end{cases} \quad (1)$$

where $P_{ned\_x}$, $P_{ned\_y}$ and $P_{ned\_z}$ represent the corresponding displacement in the local NED coordinate system. $h$ is the flight height, and $u, v,$ and $w$ are the velocity in the body axis system. The rotation matrix, translating from the NED system to the body system, is expressed as follows:

$$R_B = \begin{bmatrix} C_\theta C_\phi & C_\phi C_\psi & -S_\phi \\ S_\phi S_\theta C_\psi - C_\phi C_\psi & S_\phi S_\theta S_\psi + C_\theta C_\psi & C_\phi S_\theta \\ S_\phi C_\theta S_\psi + S_\theta S_\psi & S_\phi C_\theta C_\psi - S_\theta C_\psi & C_\phi C_\theta \end{bmatrix} \quad (2)$$

where $C_*$、$T_*$ and $S_*$ represent cosine, tangent and sine function. For the helicopter body rotation motion, the mathematic expression is:

$$[\dot{\phi},\dot{\theta},\dot{\psi}] = S_b \omega_{b/n}^b \qquad (3)$$

where $\omega_{b/n}^b$ is the angular velocity vector in the body coordinate systems, and $S_b$ is the translation matrix:

$$S_b = \begin{bmatrix} 1 & T_\theta S_\phi & T_\theta C_\phi \\ 0 & C_\phi & -S_\phi \\ 0 & C_\phi/S_\theta & C_\psi/C_\theta \end{bmatrix} \qquad (4)$$

B. *Rigid body dynamics*

Following the Newton-Euler equation, the 6-DOF rigid body dynamics of the helicopter can be derived:

$$\begin{cases} \dot{V}_b = -\omega_b \times V_b + \dfrac{F_b}{m} + \dfrac{F_{b.g}}{m} \\ \dot{\omega}_b = J^{-1}\left[ M_b - \omega_b \times J \cdot \omega_b \right] \\ J = \mathrm{diag}\left[ J_x, J_y, J_z \right] \end{cases} \qquad (5)$$

where m is the mass of the helicopter, $V_b = [u \ v \ w]$ is the velocity vector in the body frame, $\omega_{b/n}^b$ is the angular rate vector in the body frame, J is the vector of inertia moment, $F_{b.g} = [-mgS_\theta \ mgS_\phi C_\theta \ mgC_\phi C_\theta]^T$ is the projection of the helicopter's gravity vector in the body axis. $F_b$ and $M_b$ are the aerodynamic force and moment vectors:

$$F_b = \begin{bmatrix} F_x \\ F_y \\ F_z \end{bmatrix} = \begin{bmatrix} X_{mr} + X_{fus} \\ Y_{mr} + Y_{fus} + Y_{tr} \\ Z_{mr} + Z_{fus} \end{bmatrix} \qquad (6)$$

$$M_b = \begin{bmatrix} M_x \\ M_y \\ M_z \end{bmatrix} = \begin{bmatrix} L_{mr} + L_{tr} \\ M_{mr} \\ N_{mr} + N_{tr} \end{bmatrix} \qquad (7)$$

where $X_{mr}$, $Y_{mr}$, $Z_{mr}$ and $L_{mr}$, $M_{mr}$, $N_{mr}$ are the force and moment components of the main rotor; $Y_{tr}$ and $L_{tr}$, $N_{tr}$ are the force and moment components of the tail rotor; $X_{fus}$, $Y_{fus}$ and $Z_{fus}$ are the forces components of the vehicle fuselage.

The main rotor force expression is given by:

$$\begin{cases} T = (\overline{w} + a_s \overline{u} - b_s \overline{v} + \dfrac{2}{3}\Omega_{mr} R k_{col} \delta_{col} - v_{im}) \dfrac{\rho \Omega_{mr} R^2 C_{mr} b_{mr} c_{mr}}{4} \\ v_{im}^2 = \sqrt{\dfrac{v_{mr,0}^4}{4} + \dfrac{T^2}{4\rho^2 \pi^2 R^2}} - v_{mr,0}^2 \\ v_{mr,0}^2 = \overline{u} + \overline{v} + \overline{w}(\overline{w} - 2v_{im,0}) \\ \overline{V} = \begin{bmatrix} \overline{u} \\ \overline{v} \\ \overline{w} \end{bmatrix} = \begin{bmatrix} u \\ v \\ w \end{bmatrix} - V_{wind} \end{cases} \qquad (8)$$

where $\Omega_{mr}$ is the main rotor speed; $a_s$ and $b_s$ are the longitudinal and transverse tip path plane waving angles of the main rotor blade; R is the radius of the main rotor; δcol is the collective pitch steering inputs; $v_{im}$ is the main rotor induced velocity, and $v_{im,0}$ is the initial value; $k_{col}$ is the relationship gain value; $\rho$ is the air density near the main rotor; $C_{mr}$ is the rotor lift curve slope; $b_{mr}$ is the blade number; $c_{mr}$ is the chord length; $V_{wind} = [u_{wind} \ v_{wind} \ w_{wind}]^T$ denotes the wind speed along the body axis.

The force and moment components of the main rotor in Eq.6-7 can be expressed as:

$$\begin{cases} X_{mr} = -T\sin(a_s) \\ Y_{mr} = T\sin(b_s) \\ Z_{mr} = -T\cos(a_s)\cos(b_s) \\ L_{mr} = (k_\beta + Tl_{hg})\sin(b_s) \\ M_{mr} = (k_\beta + Tl_{hg})\sin(a_s) \\ N_{mr} = -(\dfrac{\rho R^2 b_{mr} c_{mr}}{14}((\Omega_{mr}R)^2 + 4.6(\overline{u}^2 + \overline{v}^2) + Tv_{im} \\ \quad + |X_{fus}\overline{u}| + |Y_{fus}\overline{v}| + |Z_{fus}(\overline{w} - v_{im})| + P_{\overline{w}}) \end{cases} \qquad (9)$$

where: c

$$P_{\overline{w}} = \begin{cases} -mg\overline{w}, & \text{if } \overline{w} < 0 \\ 0, & \text{if } \overline{w} \geq 0 \end{cases} \qquad (10)$$

and $k_\beta$ is the rotor spring constant, $l_{hg}$ is the distance between the rotor hub and the gravity center of the helicopter. The drag forces of the fuselage in Eq.6-7 can be defined as follows:

$$\begin{cases} X_{fus} = -0.5\rho S_{fux} \overline{u} l_{\max}(|\overline{u}|, |v_{im}|) \\ Y_{fus} = -0.5\rho S_{fuy} \overline{v} l_{\max}(|\overline{v}|, |v_{im}|) \\ Y_{fus} = -0.5\rho S_{fuz}(\overline{w} - v_{im})|\overline{w} - v_{im}| \end{cases} \qquad (11)$$

where $S_{fux}$, $S_{fuy}$ and $S_{fuz}$ are the effective drag area, $l_{\max}$ means taking the max value of ( ).

The forces and moments of the tail rotor in Eq.6-7 can be expressed as:

$$\begin{cases} Y_{fus} = -T_{tr} \\ L_{fus} = T_{tr}l_{htr} \\ N_{fus} = T_{tr}l_{dtr} \end{cases} \quad (12)$$

where $l_{htr}$ is the horizontal distance between the tail rotor hub and the gravity center of the helicopter, $l_{dtr}$ is the vertical distance.

*C. Main rotor flapping dynamics*

The waving dynamics of the main rotor, from cyclic pitch to the waving angle, can be described by two coupled first-order differential equations:

$$\begin{cases} \dot{a}_s = -q - \frac{1}{\tau_{mr}}a_s + A_{b_s}b_s + \frac{1}{\tau_{mr}}\theta_{a_s} \\ \dot{b}_s = -p - \frac{1}{\tau_{mr}}b_s + B_{b_s}a_s + \frac{1}{\tau_{mr}}\theta_{b_s} \end{cases} \quad (13)$$

where $\tau_{mr}$ represents the coefficient of the main rotor waving time:

$$\tau_{mr} = \frac{48R}{\gamma_{mr}\Omega_{mr}(3R-8e_{mr})} \quad (14)$$

where $\gamma_{mr}$ is the blade lock number, $e_{mr}$ is the effective hinge offset of the main rotor, $\theta_{a_s}$ and $\theta_{b_s}$ represent the longitudinal pitch angle and the lateral pitch angle:

$$\begin{cases} \theta_{a_s} = k_{lon}\delta_{lon} \\ \theta_{b_s} = k_{lat}\delta_{lat} \end{cases} \quad (15)$$

where $k_{lon}$ and $k_{lat}$ are the linkage gain value, $A_{b_s}$ and $B_{b_s}$ represent the coupling coefficient between the longitudinal waving. And the lateral waving motion can be defined as:

$$A_{b_s} = -B_{b_s} = \frac{8k_\beta}{\gamma_{mr}\Omega_{mr}^2 I_\beta} \quad (16)$$

where $I_\beta$ is the inertial moment of the rotor.

*D. Yaw channel dynamics*

The helicopter yaw response is extremely sensitive to the control signal. It is always difficult to operate for the drone operator. The control input $\delta_{ped}$ is amplified by a proportional amplifier, compared with the feedback value, which is detected by the angular rate gyroscope, generated its deviation signal, sent to the controller to calculate the tail rotor servo signal $\delta'_{ped}$. The control strategy is controlled by classical proportional-integral control law. The dynamic characteristics are as follows:

$$\delta'_{ped} = \left(k_p + \frac{k_i}{s}\right)\left(K_a\delta_{ped} - r\right) \quad (17)$$

### III. CONTROLLER DESIGN

The controller block diagram is shown in Fig.2. The attitude controller is used to ensure the helicopter Euler angles [$\phi$, $\theta$, $\psi$] and angular rate [p, q, r]. Since the position information is obtained by fusing GPS data with velocity integrated data, the dynamic characteristic of the position information is much lower than the attitude. Thus, it is safe for us to design an advanced attitude controller, simplify the designing process to realize efficient control performance.

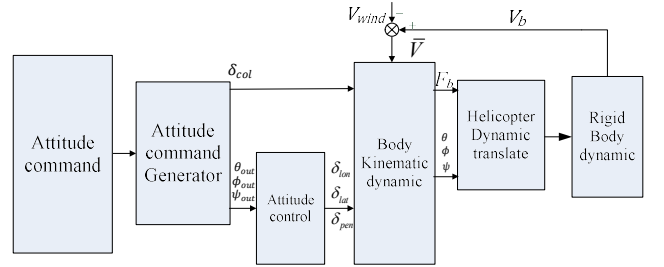

Fig.2 Structure of the attitude controller for the helicopter

*A. System Model*

The high dynamic characteristics of the attitude controller are related to the helicopter body stability. It is the key point to the entire helicopter controller. In this process, considering the wind speed in the environment with interference factor, we apply H∞ technique to minimize the disturbance effect of the wind speed.

By linearizing the model nearing the equilibrium point of the helicopter, the linearized model of the dynamic model is obtained:

$$\begin{aligned} \dot{x} &= Ax + Bu + E'V_{wind} \\ y &= y_{act} - y_{trim} \end{aligned} \quad (18)$$

where $x = x_{act} - x_{trim}$, $u = u_{act} - u_{trim}$ and $y$ are the differences between the actual state variable and the trim value. The matrix $E'$ can be obtained by injecting gusty interference into the corresponding channel then applying MATLAB system identifying toolbox. $x$, y and u are shown as below:

$$\begin{aligned} x &= \left[\phi,\theta,p,q,a_s,b_s,r,\delta'_{ped},\psi\right]^T \\ y &= \left[u,v,\phi,\theta,p,q,w,r,\psi\right]^T \\ u &= \left[\delta_{lat},\delta_{lon},\delta_{ped},\delta_{col}\right]^T \end{aligned} \quad (19)$$

*B. Attitude Controller*

The attitude controller is mainly responsible for stabilizing the attitude angle performance, so the designing of the controller does not use the collective pitch control input single. In order to ensure the system with a good attitude response, the basic output variable to be controlled is selected as $h_{out} = \begin{bmatrix} p & q & r \end{bmatrix} - h_{trim} = C_{out} x$. To guarantee the variable $h_{out}$, the following output is selected in the controller designing process:

$$h_{in} = Cx + Du \qquad (20)$$

where matrix C and D are constant matrixes to be determined by flight test.

The H∞ norm of the system transfer function can be defined as:

$$\|T_{transfer}\|_\infty = \sup l_{max}[T(jw)] = \sup \frac{\|h\|_2}{\|V_{wind}\|_2} \qquad (21)$$

The system H∞ feedback control law can be obtained as follows:

$$u = Fx + Gr_{out} \qquad (22)$$

where G and F are the system forward matrix and feedback gain matrix in the attitude controller, $r_{out} = \begin{bmatrix} \phi_{out} & \theta_{out} & \psi_{out} & \delta_{col} \end{bmatrix}$ is the signal vector generated by attitude command, which will be expressed in the next section. G and F are as follows:

$$G = -\left[C_{out}(A+BF)^{-1}B\right]^{-1}$$
$$F = -(D^T D)^{-1}(D^T C + B^T P) \qquad (23)$$

where P is semi-definite stabilizing solution of the controller Ricci equation:

$$PA + A^T P + C^T C + PEE^T P\gamma^{-2} - (PB + C^T D)(D^T D)^{-1}(D^T C + B^T P) = 0 \qquad (24)$$

The $\gamma$ expression is:

$$\gamma = \sqrt{(l_{max}(L_A R_A^{-1}))} \qquad (25)$$

where $L_A$ is the solution of the Lyapunov equation about the system matrix, $R_A$ is the solution of the Riccati equation(24).

C. *Model-free method*

In order to overcome the model-uncertain problem, a model-free method is introduced hereby to provide reference information. This method uses the position-error information to generate the attitude reference information in the model-uncertain scenario. Thus, we do not need a system identifying process in [4],[13]. The helicopter position errors are defined as:

$$\dot{E}_{xyz} = \begin{bmatrix} \dot{e}_x \\ \dot{e}_y \\ \dot{e}_z \end{bmatrix} = \begin{bmatrix} (\varepsilon_{vx} + m_x \varepsilon_{px})/\tau_x \\ (\varepsilon_{vy} + m_y \varepsilon_{py})/\tau_y \\ (\varepsilon_{vz} + m_z \varepsilon_{pz})/\tau_z \end{bmatrix} \qquad (26)$$

where $m_*$, $\tau_*$, $\varepsilon_{v*}$ and $\varepsilon_{p*}$ are the position gain value, the performance function, velocity error and position error, respectively. The values are mx=1.7, my=1.6 and mz=3.5. And the performance function is defined as:

$$\tau_i(t) = (\tau_{i,0} - \tau_{i,\infty})e^{-c_i t} + \tau_{i,\infty}, (i = x,y,z) \qquad (27)$$

where $c_i$ (i=x,y,z) is the convergence rate: cx=cy=cz=1.1. And the performance functions must satisfy the bellowing[14]:

$$\begin{cases} \tau_{i,0} > |\varepsilon_{vi} + m_i \varepsilon_{pi}|, i = x,y,z \\ \tau_{i,\infty} > 0 \end{cases} \qquad (28)$$

where $c_i$ and $\tau_{i,\infty}$ are limited to

$$\begin{cases} m_i > c_i \\ -\frac{\tau_{i,\infty}}{m_i} \leq \varepsilon_{pi} \leq \frac{\tau_{i,\infty}}{m_i}, i = x,y,z \\ -2\tau_{i,\infty} \leq \varepsilon_{vi} \leq 2\tau_{i,\infty} \end{cases} \qquad (29)$$

In the NED frame, the position errors can be expressed as:

$$\begin{bmatrix} \dot{\varepsilon}_{vx} \\ \dot{\varepsilon}_{vy} \\ \dot{\varepsilon}_{vz} \end{bmatrix} = S_b(-\omega_{b/n}^b \times V_b + \frac{F_b}{m} + \frac{F_{b.g}}{m}) - \begin{bmatrix} \ddot{P}_{ned\_x}^{ref} \\ \ddot{P}_{ned\_y}^{ref} \\ \ddot{P}_{ned\_z}^{ref} \end{bmatrix} \qquad (30)$$

Errors controller are required to designed herein:

$$u_m = \begin{bmatrix} u_x \\ u_y \\ u_z \end{bmatrix} = -KR_i \begin{bmatrix} \tanh^{-1}(e_x) + p_x \int_o^t \tanh^{-1}(e_x(T))dT \\ \tanh^{-1}(e_y) + p_y \int_o^t \tanh^{-1}(e_y(T))dT \\ \tanh^{-1}(e_z) + p_z \int_o^t \tanh^{-1}(e_z(T))dT \end{bmatrix} \qquad (31)$$

where $K = diag(\frac{k_x}{(1-e_x^2)\tau_x(t)}, \frac{k_y}{(1-e_y^2)\tau_y(t)}, \frac{k_z}{(1-e_z^2)\tau_z(t)})$, $k_i$ and $p_i$ (i=x,y,z) is positive gain value: kx=0.16, ky=0.13, kz=0.06, px=0.3, py=0.4 and pz=0.7.

After transforming the intermediate um into spherical coordinates, the intermediate command in Eq. 22 $r_{out} = \begin{bmatrix} \phi_{out} & \theta_{out} & \psi_{out} & \delta_{col} \end{bmatrix}^T$ can be calculated:

$$\begin{cases} \phi_{out} = -\sin^{-1}(\dfrac{u_y}{\|u_m\|}) \\ \theta_{out} = -\tan^{-1}(\dfrac{u_x}{u_z}) \\ \psi_{out} = \psi^{ref} \\ \delta_{col} = -\|u_m\| \end{cases} \quad (32)$$

## IV. EXPERIMENT

*Platform*

To verify the efficiency and robustness of the proposed method, a flight test is executed on a helicopter platform, as shown in Fig.3. The model parameters of the helicopter are obtained by measuring manually and the main parameters are shown in Table 1. The small size helicopter is equipped with the onboard ARM, IMU, and real-time kinematic (RTK) GPS. The measurement accuracy of the velocity is 5cm/s, the attitude is 0.1 degree, and the position is 5cm. An onboard storage system is designed to record flight data in memory card.

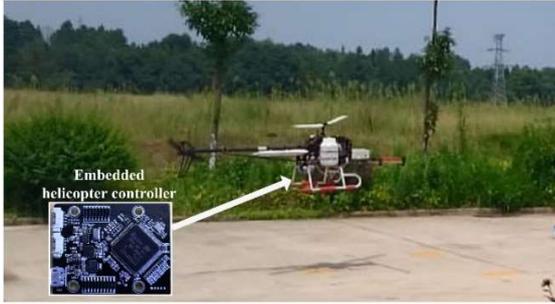

Fig.3 Experimental helicopter platform

*Experiment Results*

The helicopter takes off from a platform with $P_{ned\_x}=0$ m, $P_{ned\_y}=0$ m, $P_{ned\_z}=0.20$ m, $\psi=0$ degrees, flies to the pre-set hovering point, rises to $P_{ned\_z}=0.65$ m, turns heading angle to $\psi=273.5$ degrees, hovers over the ground for about 20 minutes, then lands on the ground finally. The flight path is shown in Fig.7. In this process, an adjustable speed fan is set beside the area to provide wind guest with 3m/s pointing in the direction as shown in Fig.7.

The attitude performances are shown in Fig.4. It's can be obtained that the pitch and roll channels have great coupling induction. The coupling effect gets stronger, especially under the wind guest and ground effect phenomena conditions when the platform flies near the ground. It can be observed how highly-coupled the system is during the take-off process. While the proposed controller can still control the helicopter more smooth and less overshoot than the method does in [13]. The yaw channel realizes accurate tracking performance because the yaw channel is independent. After leaving the ground, the helicopter then hovers over the set point in Fig.7. The attitude responses of the helicopter show more stable than its performance during take-off process in Fig.4. Also, the proposed method shows more efficiency and robust ability, has little fluctuation than the method does in [13]. The angle rates, shown in Fig.5, can also demonstrate it. The proposed controller outputs set in Eq. (32) are shown in Fig.6.

**Table 1**
Main variable values of the helicopter

| Parameter | Value (unit) | Parameter | Value (unit) |
|---|---|---|---|
| $m$ | 7.6kg | $R$ | 0.82m |
| $J_x$ | 0.19kg.m² | $k_{lat}$ | 0.53 |
| $J_y$ | 0.46 kg.m² | $k_{lon}$ | 0.54 |
| $J_z$ | 0.31 kg.m² | $k_{col}$ | 3.77 |
| $g$ | 9.81N/kg | $c_{mr}$ | 0.06m |
| $\gamma_{mr}$ | 1.131 | $b_{mr}$ | 2 |
| $\Omega_{mr}$ | 175.2rad/s | $I_\beta$ | 0.0913kg.m² |

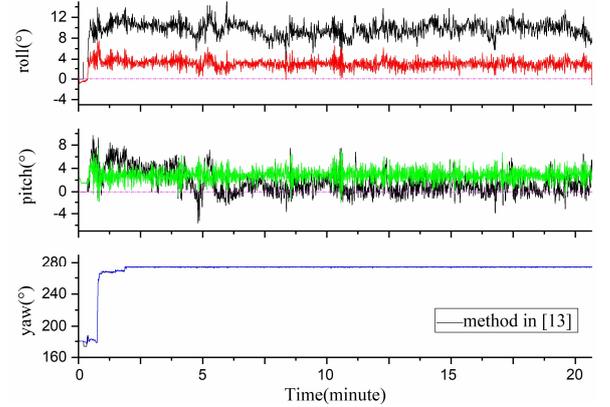

Fig.4 the Attitude responses of the vehicle

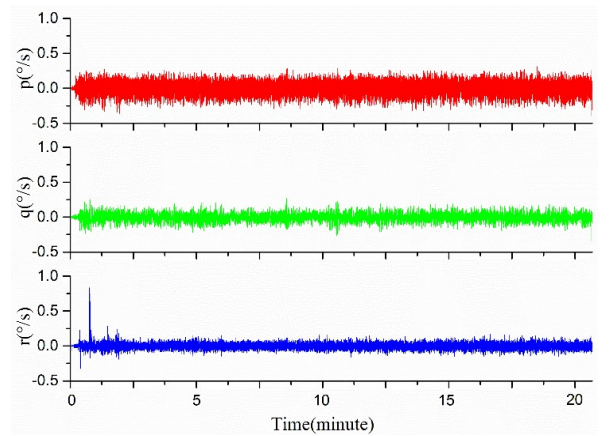

Fig.5 the Angle rate responses of the vehicle

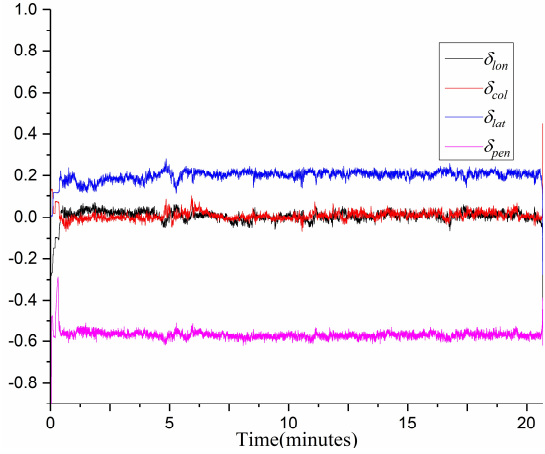

Fig.6 the Outputs of the proposed controller in Eq. (32)

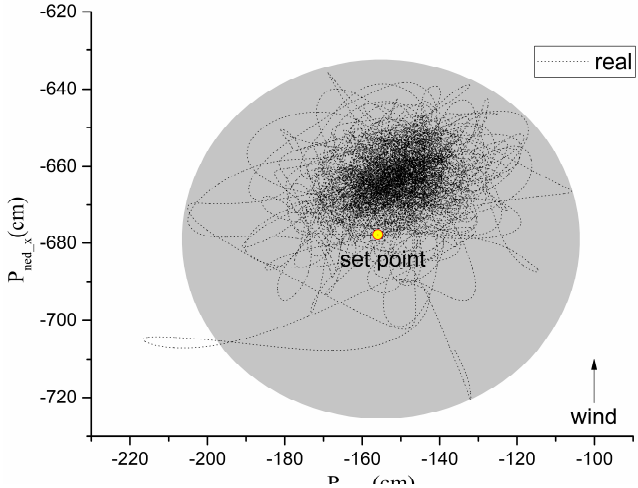

Fig. 7 Hovering flight path

## V. CONCLUSION

A robust attitude control method was proposed for a small size helicopter. A series of attitude performance results are given to demonstrate the effectiveness and robustness of the proposed control algorithm. However, only taking off, hovering and landing scenarios are involved in this paper. In the future, some agility trajectory tracking with high-speed tests and advanced position tracking controller design can be done.